\documentclass[conference]{IEEEtran}
\usepackage{graphicx}
\usepackage{listings}
\usepackage{xcolor}
\usepackage{amssymb}
\usepackage{amsmath}

\begin{document}

\title{Methods for Accountability in Machine Learning Inference Services}

\makeatletter

\makeatother
\author{\IEEEauthorblockN{
Mustafa Canim*,
Ashish Kundu*,
Joshua Payne*
}

\IEEEauthorblockA{IBM Thomas J. Watson Research Center, Yorktown Heights, NY, USA.\\ \{akundu,mustafa\}@us.ibm.com, joshua.f.payne@ibm.com}
*Authors contributed equally to this work.
}

\maketitle

\begin{abstract}
Classification-as-a-Service (CaaS) is widely deployed today in machine intelligence stacks for a vastly diverse set of applications including anything from medical prognosis to computer vision tasks to natural language processing to identity fraud detection. The compute power required for training complex models on large datasets to solve these problems can be very resource-intensive. The CaaS model may cheat by fraudulently bypassing expensive training procedures in favor of weaker, less computationally-intensive algorithms which yield results of reduced quality. Our work addresses the following questions, given a classification service supplier $S$, intermediary CaaS provider $P$ claiming to use $S$ as a classification backend, and client $C$: (i) how can $P$'s claim to be using $S$ be verified by $C$? (ii) how might $S$ make performance guarantees that may be verified by $C$? and (iii) how might one design a decentralized system that incentivizes service proofing and accountability? To this end, we propose a variety of methods for $C$ to evaluate the service claims made by $P$ using probabilistic performance metrics, instance seeding, and steganography. We also propose a method of measuring the robustness of a model using a blackbox adversarial procedure, which may then be used as a benchmark or comparison to a claim made by $S$. Finally, we propose the design of a smart contract-based decentralized system that incentivizes service accountability to serve as a trusted Quality of Service (QoS) auditor.

\end{abstract}


\IEEEpeerreviewmaketitle

\section{Introduction}\label{intro}

The market for Machine Learning as as Service (MLaaS) has been exploding recently\footnote{https://www.marketsandmarkets.com/Market-Reports/deep-learning-market-107369271.html}, with breakthroughs in techniques, availability of datasets, demand for machine-intelligence solutions, and power of computation rapidly increasing. With this, we've seen a massive demand for resources to support such computation, and these resources aren't cheap. Indeed, much of machine learning commercialized today relies on deep learning models trained and ran on large GPU clusters, requiring a great deal of time, energy, and memory to train. 

From a business perspective, it is optimal for the service entity to minimize costs while maximizing revenue. However, when considering ways to minimize costs, a Classification-as-a-Service (CaaS) supplier may be tempted to cheat by delivering cheaper services, classifications made by less computationally-intensive models, while charging for and promising classification services performed by better models that require greater resources. A classification service supplier exposes an API that takes as input an image and responds with a softmax probability distribution over the possible labels. The supplier claims to be using a deep neural network model with hundreds of layers trained on Tencent ML-Images\footnote{https://github.com/Tencent/tencent-ml-images} but instead uses a Support Vector Machine (SVM) model trained on Cifar-10\footnote{https://www.cs.toronto.edu/~kriz/cifar.html}. SVMs are typically less computationally intensive than deep learning models, but also often do not perform as well, especially on computer vision tasks. The overall effect is a sub-standard level of quality, where the service accuracy, efficiency, speed, and/or robustness is below the standard set by the Service Level Agreements (SLA) guarantees or QoS claims that a CaaS provider makes.

Likewise, intermediary providers claiming to use a high-quality, but expensive, service as the classification backend may be tempted to instead use a cheaper or homegrown classification service. For example, Google's Bidirectional Encoder Representations from Transformers (BERT) \cite{bert} model for natural language processing (NLP) is used as a pretraining procedure, generating 768-dimensional embeddings for input text. Predictions using these high-dimensional embeddings are often of high quality, but can be more expensive to generate and train with. A classification provider claims to use BERT via Google's NLP API in the backend for pretraining to classify input text, but instead uses a homegrown model to generate 50-dimensional embeddings. The classifications resulting from models with this pretraining procedure will likely be of sub-standard quality.

The aim of this paper is to define these and similar problems that may arise more formally and to propose methods that address each of them.








\begin{figure}[t]
    \centering
    \includegraphics[width=9cm]{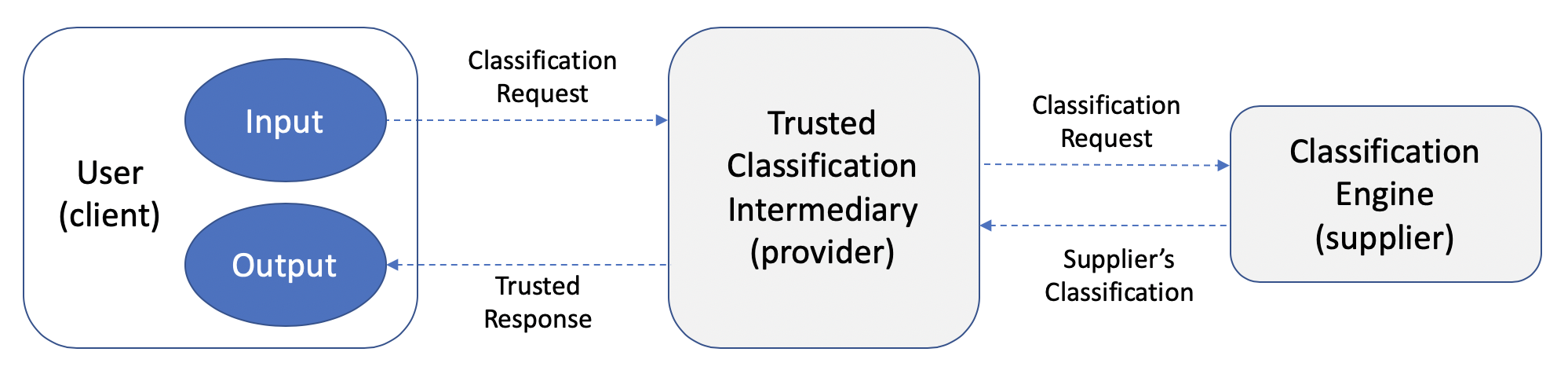}
    \caption{A trusted intermediary $P$ queries the promised supplier $S$ when a request from $C$ is received.}
    \label{fig:problem_trusted}
\end{figure}
\section{Problem Definition}\label{problem}
We define a classification service supplier $S$ to be a service entity that exposes an API that uses a model $M$ that, when trained, computes a function $M_t: X\rightarrow \hat{Y}$ which takes as input an input object $x\in X$ (e.g., a sequence of DNA, an image, a sequence of text, an identity graph) and returns as output a classification vector $\hat{y}\in \hat{Y}$, in which $\hat{y}_i$ corresponds to the model's probability estimate of $x$ being an instance of object class $i$, which is described by the ground truth label vector $y$. We define a classification service provider $P$ to be an intermediary entity that takes as input a classification query $x$, performs some process, and returns as output a probability vector $\hat{y}$. If $P$ makes an accurate claim to be sourcing classification service from $S$, then the process performed would be sending $x$ along to $S$ and defining $\hat{y}$ to be the output received from $S$. For now, we put aside scenarios where some transformation is applied by $P$ to the input or output, as these would likely require case-by-case solutions. Finally, we define a client $C$ to be an entity with an object $x$ that it wants a classification vector $\hat{y}$ for.

There are many scenarios in which validating the QoS of a classification service supplier or provider becomes important. For the following scenarios, assume $M$ is untrained.\\
\begin{figure}[t]
    \centering
    \includegraphics[width=9cm]{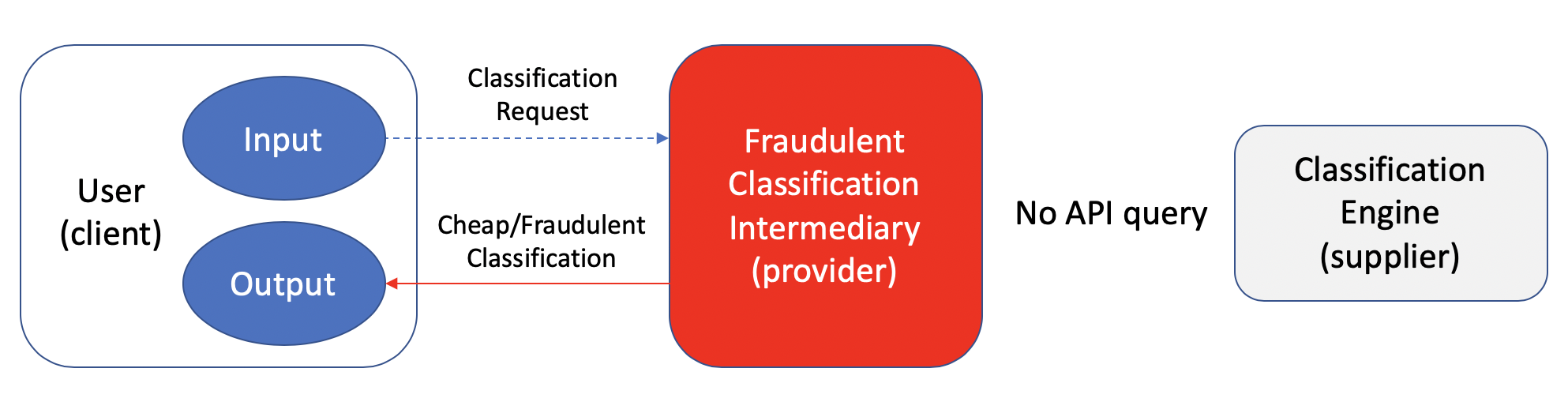}
    \caption{An untrusted, cheating intermediary $P$ does not query the promised supplier $S$ when a request from $C$ is received, possibly returning cheaper classification results.}
    \label{fig:problem_cheating}
\end{figure}
\textit{Scenario 1: S trusted, P untrusted, M blackbox}\\ \\
In this scenario, $C$ requests classification $y$ of object $x$ from $P$, who in turn claims to use a service $S$ as a classification backend. $C$ also does not have access to $M$. However, we consider several scenarios here with different types of interaction with $S$. One scenario we consider is one where $M$ is capable of steganography and $C$ can in turn perform reasonable computations on its own. Another scenario is where $C$ has the ability to interact directly with $S$ and an instance of $M$. This again may be broken into two scenarios, one where some modifications or features $M$ may be added, and one where that is not the case. A final scenario is one where $S$ quality-proofs $y$ by accompanying the output with metadata, such as a certificate or explanation of the classification.\\ 

\textit{Scenario 2: S trusted, P untrusted, M whitebox}\\ \\
Here, $C$ again requests classification $y$ of object $x$ from $P$, but this time, the model is known. We now split this into two scenarios. In the first, we assume that $M_t$ is reasonably computable by $C$. In the second, we assume that $M_t$ is not reasonably computable by $C$, but that a decentralized system may compute it and communicate with $C$.\\

\textit{Scenario 3: S untrusted, M blackbox}\\ \\
Now we consider a scenario in which $S$ makes a performance guarantee, but is untrusted. Hence, $C$ (which may be $P$, in this case) must be able to verify these performance guarantees. In this scenario, $M$ is unknown, and we consider performance metrics of accuracy, robustness, and latency. In particular, we're concerned with the absolute metrics as well as their consistency with respect to $C$.

\section{Related Work}
Machine Learning-as-a-Service (MLaaS) is a subset of cloud computing, for which Service Level Agreements and Quality of Service measures have been well-studied. The authors of \cite{cloud_b} have proposed a framework for considering the technological, legal, consumer-adoption, and infrastructure security problems that arise in cloud computing, proposing that these issues necessitate research into topics such as optimal risk transfer and SLA contract design. We take this as a call-to-action, as we feel that SLAs for MLaaS applications can be better designed with respect to the possibility of intermediary fraud. The authors of \cite{wsla} define web service level agreements formally with three entities: parties, which describe the service provider, the service consumer, and third parties; SLA parameters, which describe resource metrics and composite metrics, which provide insightful and contextual information where raw resource metrics fall short, and Service Level Objectives (SLOs), which are a set of formal \textit{if: then} expressions. We acknowledge the necessity for this formal, deterministic language in our context as well when it is applicable, but also note that in some cases where only probabilistic measures are available for QoS, the SLAs which address these must be more expressive to accommodate. We also note that, for instance, Google's ML Engine SLA only covers metrics such as uptime, and not metrics related to model accuracy or robustness\footnote{https://cloud.google.com/ml-engine/sla}.
Our work seeks to identify methods for determining whether some promise in a MLaaS provider's SLA or other guarantee mechanism is fraudulent. The first method we describe uses the ideas of steganography in the deep learning domain, proposed by \cite{deep_steg}, to embed hidden messages into an input request that a supplier would correctly identify while a fraudulent provider would not be able to identify it. This work draws inspiration from steganographic methods for images that do not necessarily involve deep neural networks \cite{steg} as well as autoencoders for dimensionality reduction \cite{autoenc} to propose a deep neural network architecture that embeds a message image into a cover image such that the message image is not visible to a human eye, and can be extracted by a separate component of the network.
Where supplier services offer explanations with their predictions, we propose the use of explainability rating for our service access framework. The author of \cite{xai} gives a clear overview of the challenges and current approaches to explainability and interpretability in machine learning, which we use. 
Supplier services also have limitations when it comes to robustness, and these limitations can be indicative of the robustness quality of the service provided. To study this, we reference the approach proposed in \cite{gans} to generate adversarial examples with respect to different classes offered for classification by the supplier.
In these models, we also utilize seeding capabilities offered by many machine learning libraries and frameworks such as Tensorflow \cite{tf}, PyTorch \cite{pytorch}, and Caffe \cite{caffe}. These seeding methods allow for reproducibility, giving the client or a federated accountability entity the ability to verify claims being made by MLaaS providers.
In proposing a federated accountability entity, we build on the work by the authors of \cite{hyperledger} and \cite{permissioned}---we conjecture that a permissioned blockchain with peers which verify model outputs and reach consensus on the validity of the output of a provider is a well-suited model for a federated accountability entity. 

\begin{figure*}[t]
    \centering
    \includegraphics[width=15cm]{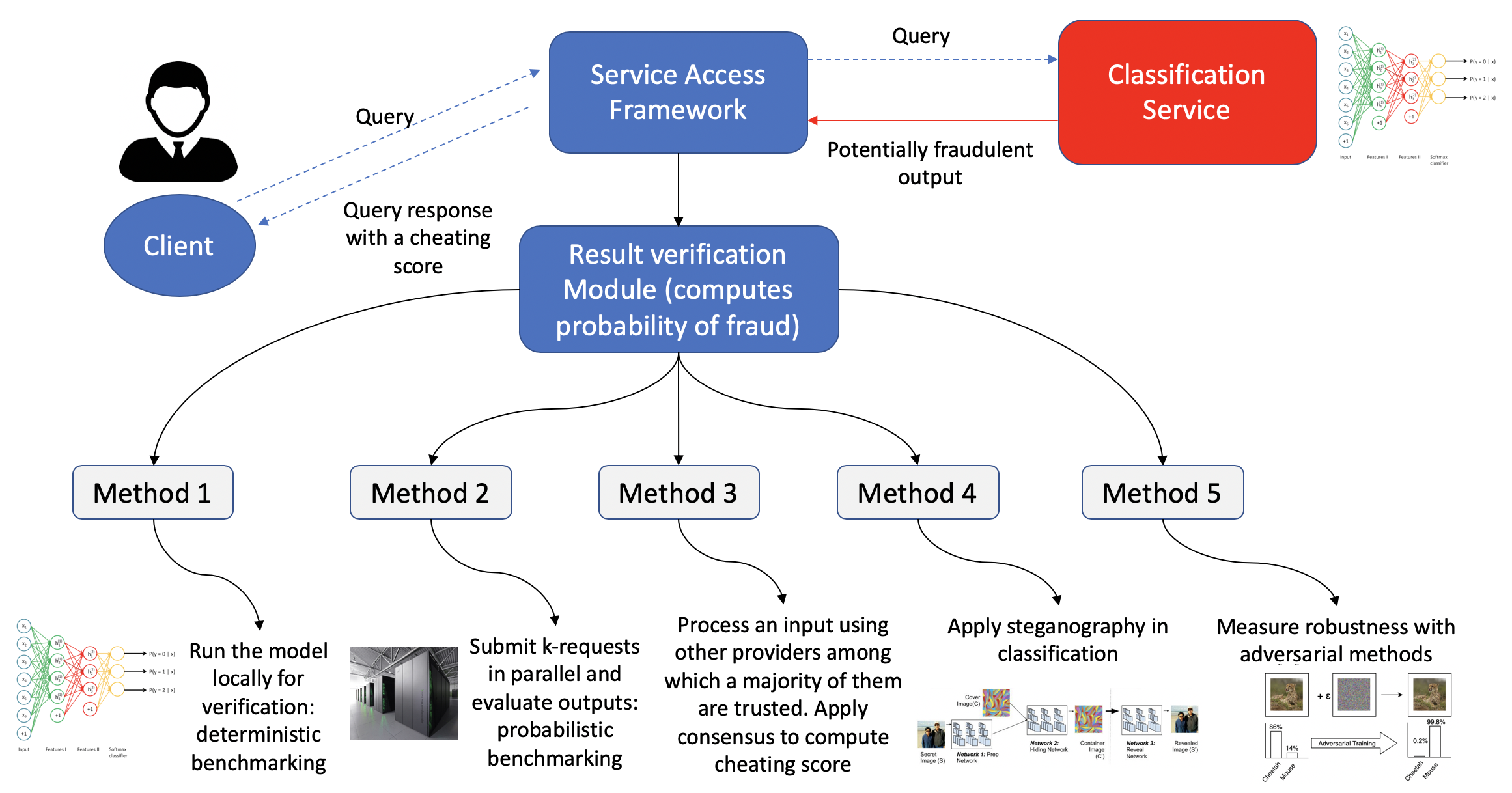}
    \caption{We propose as a solution a set of five main methods described in section III.}
    \label{fig:solution}
\end{figure*}
\section{Proposed Solutions}\label{solution}
In this section, we present an overview of our solutions to the scenarios described in \ref{problem}.

\subsection{Watermarking, Seeding, Benchmarking, and Explainability}
In Scenario 1, we considered the case where the CaaS supplier $S$ is trusted, but $P$ is untrusted. Likewise, the supplier's model $M$ is blackbox; that is, the model's architecture and hyperparameters are unknown. Within this scenarios, we identified several sub-scenarios that will allow for solutions involving different techniques and approaches. \\
\subsubsection{Deep Steganography}
For our first sub-scenario, we consider the case where $S$ employs a model that supports deep steganography. Steganography, a term that dates back to 1499\footnote{https://en.wikipedia.org/wiki/Steganography}, is the practice of concealing a message using a medium. Examples of steganographic messages are those written on envelopes in the area covered by postage stamps, or those written within the lowest bits of noisy images. Unfortunately, typical algorithmic steganography techniques fail here, since any steganographic protocol introduced by $S$ intended to be made known to $C$ is also known by $P$. In order to address this, we build on a body of work on deep stegangraphy proposed by Shumeet Baluja in \cite{deep_steg}. In this model, which takes an approach similar to auto-encoding networks, the hidden message is nearly impossible to detect and decode without explicit knowledge of the trained revealing/classification model $M_t$. An outline of the procedure is as follows:
\begin{enumerate}
    \item $S$ creates a steganographic service that takes as input $x$ and returns as output $\hat{y}_i$, where $\hat{y}_1,\ldots,\hat{y}_n$ are classification probabilities for $n$ object classes, and $\hat{y}_{n+1}$ is the classification probability for the message class.
    \item $S$ defines a procedure for creating a model $m$ that will take as input an instance of any object class and return as output an instance of the message class that ``looks" like the original object class---that is, $P$ would not classify it as the message class with high probability. This procedure can be likened to the generative adversarial models described by Goodfellow et al. in \cite{gans}.
    \item $C$ leverages $m$ to generate instances of steganographic objects and queries $P$ with some random sequence of steganographic objects and nonsteganographic objects.
    \item If the classification results $\hat{y}$ for the steganographic objects are mostly correct, it may be deemed that $P$ is likely not fraudulent in its claim. However, if the classification results $\hat{y}$ for the steganographic objects are mostly incorrect, it may be deemed that $P$ is likely fraudulent and not using $S$.\\
\end{enumerate}

\begin{figure}[t]
    \centering
    \includegraphics[width=3.35in]{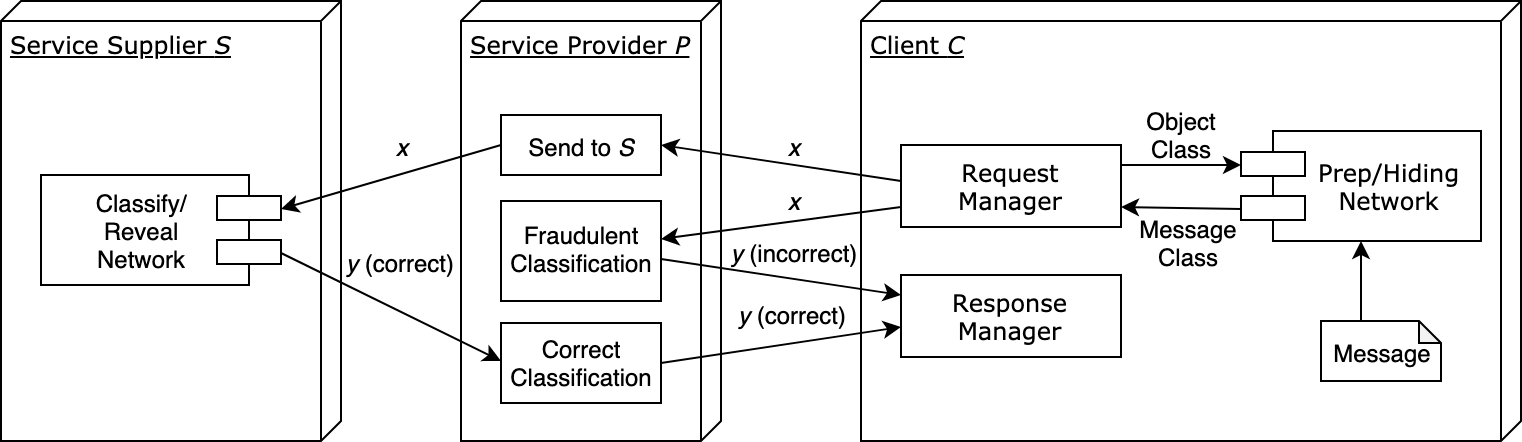}
    \caption{The flow of our proposed steganography-based MLaaS accountability testing method.}
    \label{fig:steg}
\end{figure}

The deep steganography model shown in \ref{fig:steg} has three components, though they are trained as one model: the prep network, the hiding network, and the revealing network, which performs classification. 

The prep network is responsible for distributing the bits of the message class instance among the object class instance, or ``cover", as well as efficiently encoding the message class instance. The hiding network takes as input the output of the prep network, which is the message class instance, processed for embedding into the cover, as well as the cover itself. This network outputs an instance of a ``container" which is the desired steganographic object. $C$ is responsible for providing a cover and message, as well as running these two networks. The reveal network is run by $S$ and takes as input the hiding network and decodes it to reveal the hidden message. The overall model implements the loss function $$ \mathcal{L}(c, c', s, s') = ||c - c'|| + \beta||s - s'||, $$ where $c$ is the cover instance, $c'$ is the container instance, $s$ is the secret message, $s'$ is the reconstructed message, and $\beta$ is a hyperparameter used to weigh reconstruction errors. Intuitively, this penalizes large distances between the input cover and the container instances, as well as large distances between the secret message and the reconstruction. $L_n$-norm loss functions for different choices of $n$ could be used as well. As these networks are trained as one model, $S$ is intended to instantiate and train the full model and publicly release the prep network and hiding network for use by $C$, as well as instances of the message class (which could conceivably be extended to multiple classes). 
In describing this procedure, we've assumed that generating a cheaper model that is not $M_t$ and that can discriminate between inputs of the message class and object class is not viable.\\

\subsubsection{Deterministic Benchmarking} 
For this sub-scenario, we consider the case where $S$ makes use of seeding. A seed is a number or vector used to initialize a pseudorandom number generator. Hence, any pseudostochastic process (such as stochastic gradient descent) or pseudorandom generation (such as weight initialization) used in machine learning models can be made reproducible by seeding. As a result, we propose that a supplier $S$ make available the seed configuration used to initialize $M$ for any provider $P$, so that a client $C$ can initialize another instance directly with $S$ using the same seed configuration and compare $(x, \hat{y})$ pairs yielded from $P$ and $S$. We have determinism due to seeding, so if each pair is identical, we can say with high probability that $P$ is not fraudulent. On the other hand, if any pair is not identical, we can say that $P$ is fraudulent. 

\subsubsection{Probabilistic Benchmarking} \label{pbench}
Now we consider the case where $S$, while trusted, has a strictly blackbox model. Because we're unable to compare $(x, y)$ pairs exactly, we must settle for a more probabilistic approach of performance benchmarking. As an example, we consider accuracy as the performance metric under scrutiny. $C$ prepares $k$ pairs of identical inputs $(x_i, x_i)$ for which it has ground truth labels, sends $P$ and $S$ each one of each $x$, retrieves the outputs $(y_{1_i}, y_{2_i})$, and calculates the overall accuracy for each model. If $P$ is shown to have lower accuracy than $S$ (this may be evaluated with variance in mind, which is controlled by $k$), it is likely that $P$ is fraudulent. If they show roughly the same accuracy (again considered with respect to the size of $k$), then it's likely that $P$ is not fraudulent.\\

\subsubsection{Metaresults} 
Finally, we consider the case where $S$ may provide information along with the results of a classification, such as an explanation of the results or a ``quality assurance" key that may be used to verify a classification event with $S$. In this case, $C$ would need to know of the quality assurance measures taken by $S$ in order to take advantage of them. This being the case, it would be very difficult for an untrusted provider $P$ to forge the metaresults returned from a verified classification event with $S$. Likewise if $P$ indeed queried $S$, all it would need to do is pass the metaresults along to $C$. This method is predicated on the assumption that it is difficult for any provider $P$ without access to $M_t$ to create valid metaresults.\\

\subsection{Accountability with Known M} 
Let's now consider the scenario in which $M$ is a whitebox model. That is, $S$ has made publicly known the architecture, hyperparameters, and training procedure. Here, we have two subscenarios. In the first, we assume that $M_t$ is also available or reasonably computable by $C$. In this case, note that $C$ can use the performance metric comparison method described in \ref{pbench} by instantiating its own $M_t$ and comparing the metrics given by identical trials run with $P$ and its own model. 

On the other hand, we consider a scenario in which $M_t$ is not available or reasonably computable by $C$. This particular scenario is one of the more common scenarios in today's CaaS market, along with blackbox scenarios, as the service suppiers often release research papers and open-source code detailing their work. Here, we introduce a decentralized system that functions as a trusted QoS auditor. The QoS auditor is a blockchain-based model with peers being CaaS accounters, which are oracles who compute $M$ and post transactions with data including performance metrics of the model and performance metrics of $S$. Majority consensus mechanisms here are designed to ensure that accurate performance metrics of a service $S$ are maintained on-chain. Clients who desire a verification of a model $M$ send requests with tokens to the chain to receive verified performance metrics on $M$, and oracles that cast votes on performance metrics of $M$ which are in the majority are rewarded with tokens. The blockchain is not open to the public, since the services provided to the client are more valuable than the tokens the client pays for verification. Hence, view access would be managed by a trusted gatekeeper. An assumption we make here is that the oracles are trusted and/or mutually untrusting, and would therefore not collude to cheat the system.
\begin{figure}[t]
    \centering
    \includegraphics[width=3.35in]{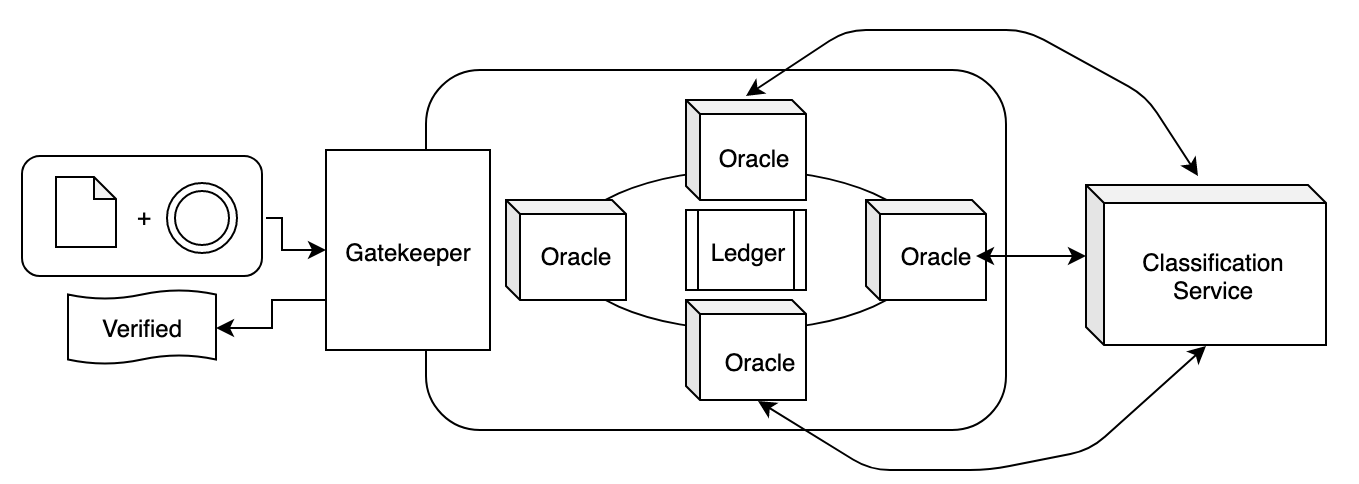}
    \caption{An outline of a decentralized MLaaS service accountability system based on the permissioned blockchain model.}
    \label{fig:ledger}
\end{figure}
\subsection{Benchmarking Performance Claims of Untrusted S}
\begin{figure}[t]
    \centering
    \includegraphics[width=3.35in]{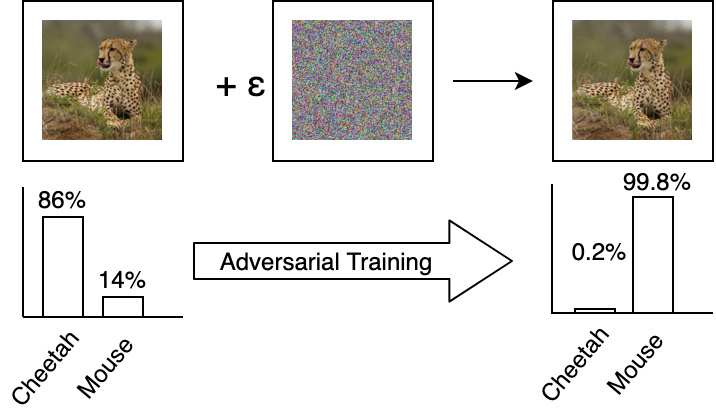}
    \caption{A visual representation of the process by which adversarial inputs are generated.}
    \label{fig:adv1}
\end{figure}

\begin{figure}[t]
    \centering
    \includegraphics[width=3.35in]{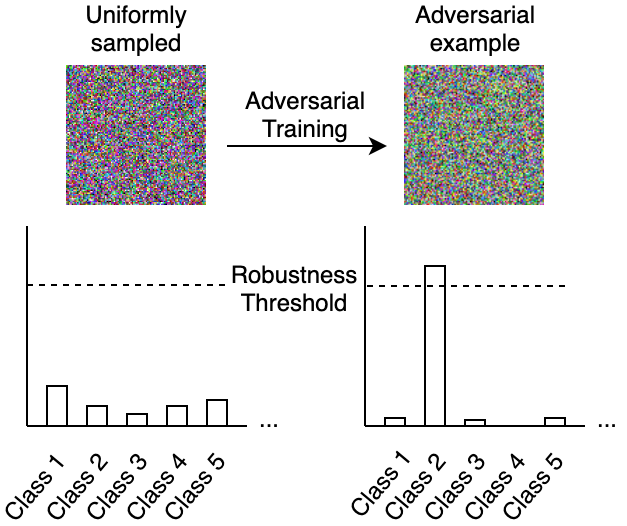}
    \caption{One way of testing the robustness of a classification service is to study the gradient of the outputs during adversarial training, with respect to each class.}
    \label{fig:adv2}
\end{figure}

Robustness is a desirable characteristic in classification models. Recent work by Ian Goodfellow and others explores the methods and implications of real world and computationally-generated adversarial examples against blackbox models, such as $M$ \cite{gans2}\cite{gans3}. These models take advantage of the fact that the function $M_t$ is differentiable and by studying the gradient with respect to the input, the output can be controlled in some sense. The canonical multilayer perceptron example from Goodfellow's 2014 paper between an adversarial network and a discriminator network illustrates this. Suppose G is a differentiable function that represents the generator multilayer perceptron. In particular, if $\theta_g$ is the set of model parameters for $G$, $p_z(z)$ is the prior probability distribution of the input $z$, then let $G(z,\theta_g)$ map to the data space $X$, described by a multilayer perceptron. This model wants to learn the generated distribution, $p_g$, over the data. Now, if $\theta_d$ is the set of model parameters for $D$ and $x\in X$, let $D$ map to the space of real scalars, and note that $D$ is also described by a multilayer perceptron. Then $D(x,\theta_d)$ represents the probability that $x$ came from the data, and not $p_g$. We then formulate the minimax game to represent the objective, with respect to the value function $V(G, D)$, where $p_t$ is the true probability distribution of the data: \begin{align*}
\min_G\max_D V(D, G) =& \mathbb{E}_{x\sim p_{t}(x)}[\log D(x)]\\
&+ \mathbb{E}_{z\sim p_z(z)}[\log(1-D(G(z)))].
\end{align*}
For our purposes, we are n  concerned only with a dynamic generator model, where the discriminator model does not respond to adversarial adaptation. Hence, we want 
\begin{align*}
\arg\max_G \left(\mathbb{E}_{x\sim p_{t}}[1-D(x)]+ \mathbb{E}_{x\sim p_g}[D(x)]\right).
\end{align*}
where the sum being maximized is the expected absolute error of the discriminator. To illustrate the idea behind generative adversarial models for a robustness metric, consider figure \ref{fig:adv1}. \\
Here, $M$ is a binary classification model that is tasked with assigning probabilities of an input example being labeled as ``cheetah" (class 1) or ``mouse" (class 2). The model correctly predicts the label of ``cheetah" for the input example above. However, we introduce an adversarial model for $M$, $A_M$, which uses the gradient of the outputs of $M_t$ to generate an additive transformation (seen in the center panel) which, when applied to the input, yields an output that is classified as ``mouse" with high probability, though this adversarial example is clearly, to the human eye, still depicting a cheetah. To establish a measure of robustness, we concern ourselves with the gradient of convergence to a certain robustness threshold given an input image that is uniformly sampled from the space of all input images. For a simple guarantee metric here, we look at the average number of input queries the adversarial model needs to make to the blackbox model before we reach a certain classification probability for class $i$, from a uniformly sampled image. In the figure \ref{fig:adv2}, we see that an adversarial model has generated an example that has passed the robustness threshold for class 2.\\
In this scenario, $S$ may release the architecture for an adversarial model $A_M$, as well as a robustness score (perhaps a sigmoidal function of the average number of examples needed for $A_M$ to generate before reaching a certain robustness threshold) for each class that the service offers classification for. This need not be limited to images, but can be extended to any type of input.


\section{Conclusion}
In this paper, we've examined the problem of Machine Learning-as-a-Service (MLaaS) accountability and have proposed a suite of accountability methods that may be used to verify the Quality of Service (QoS) guaranteed by a MLaaS provider in their Service Level Agreements (SLAs) or otherwise. These methods build upon techniques in deep steganography, seeding for reproducibility in models, explainability in artificial intelligence (XAI), generative adversarial models for robustness measurements, and permissioned blockchains for federated accountability entities. With these methods, a client will be able to determine a probability of whether a MLaaS provider is cheating, and MLaaS providers will be able to use these methods to create improved SLAs that reflect respective verifiable QoS measures.

\bibliographystyle{IEEEtran}
\bibliography{IEEEabrv,references}

\end{document}